\documentclass{article}
\usepackage[utf8]{inputenc}
\usepackage{textgreek}
\usepackage{spconf,amsmath,graphicx,booktabs}

\usepackage{enumitem}
\usepackage{caption}
\setlist{nosep, leftmargin=14pt}

\usepackage{mwe} 

\newcommand\blfootnote[1]{%
  \begingroup
  \renewcommand\thefootnote{}\footnote{#1}%
  \addtocounter{footnote}{-1}%
  \endgroup
}
\makeatletter
\renewcommand{\footnoterule}{%
  \kern-3pt
  \hrule width \columnwidth height 0.4pt
  \kern 2.6pt
}
\makeatother
\makeatother

\title{Evaluating TabPFN for Mild Cognitive Impairment to Alzheimer's Disease Conversion in Data-Limited Settings}
\name{\begin{tabular}{c}
Brad Ye$^{1*}$ \qquad Bulent Soykan$^{2}$ \qquad Gulsah Hancerliogullari Koksalmis$^{3,4}$ \\
Hsin-Hsiung Huang$^{5}$ \qquad Laura J. Brattain$^{1}$
\end{tabular}}

\address{$^{1}$ Department of Medicine, University of Central Florida College of Medicine, Orlando, FL, USA \\
$^{2}$ Department of Mechanical, Industrial \& Manufacturing Eng., The University of Toledo, Toledo, OH, USA \\
$^{3}$ Department of Industrial Eng. and Mngt. Systems, University of Central Florida, Orlando, FL, USA \\
$^{4}$ Department of Industrial Engineering, Istanbul Technical University, Istanbul, Turkey\\
$^{5}$ School of Data, Mathematical, and Statistical Sciences, University of Central Florida, Orlando, FL, USA}

\begin{document}
\maketitle
\blfootnote{$^{*}$Brad Ye is a volunteer researcher at the NIMBUS Lab, University of Central Florida.}

\begin{abstract}
Accurate prediction of conversion from mild cognitive impairment (MCI) to Alzheimer's disease (AD) is important for earlier risk stratification and intervention, but model development is constrained by limited longitudinal follow-up and incomplete biomarker data. We evaluate Tabular Prior-data Fitted Networks (TabPFN) against standard machine learning baselines for 3-year MCI-to-AD conversion prediction using TADPOLE, a dataset derived from the Alzheimer's Disease Neuroimaging Initiative (ADNI). Biomarker-only features included demographics, apolipoprotein E $\epsilon$4 (APOE $\epsilon$4) status, structural magnetic resonance imaging (MRI) volumetrics, and positron emission tomography (PET) imaging; cognitive scores directly tied to diagnostic criteria were excluded to reduce information leakage. We compared TabPFN with logistic regression, random forest, LightGBM, and tuned XGBoost across training sizes of $N=50$ to $N=500$. TabPFN achieved an area under the receiver operating characteristic curve (AUC) of 0.879 on the test evaluation set, closely tracking tuned XGBoost (AUC=0.887) and above LightGBM (AUC=0.886) and exceeding logistic regression (AUC=0.858). In the smallest training setting ($N=50$), TabPFN obtained the highest AUC (0.810). However, TabPFN's balanced classification accuracy (BCA) was poor at the default 0.5 threshold but improved after threshold optimization, indicating that its outputs should be treated as relative risk scores unless calibrated. These results suggest that tabular foundation models may be useful for biomarker-based AD risk prediction in data-limited settings, provided that decision thresholds are validated on representative cohorts.
\end{abstract}

\section{Introduction}
\label{sec:intro}
The number of Americans living with Alzheimer's disease (AD) is projected to reach 13.8 million by 2060 \cite{alzassoc2024facts}, underscoring the urgent need for improved early detection and intervention strategies. Machine learning models show promise for predicting disease progression, yet their development faces a fundamental challenge: limited high-quality training data \cite{2025StatisticalLearning}. 

Alzheimer's datasets such as Open Access Series of Imaging Studies (OASIS) and Alzheimer's Disease Neuroimaging Initiative (ADNI) suffer from class imbalances, missing longitudinal measurements, and incomplete patient histories—constraints that significantly reduce usable training samples \cite{ai2025mci}. This limitation is particularly acute for slowly progressing diseases like AD, where collecting sufficient longitudinal observations with complete neuroimaging and biomarker profiles requires years or decades. Recent work has explored several strategies for addressing these data scarcity challenges in AD modeling, including generative augmentation via diffusion models \cite{GonzalezNunez2025} and multi-modal digital twin frameworks that integrate uncertainty quantification \cite{soykan2026cognitivetwinrobustmultimodaldigital,soykan2026personalizeddigitaltwinscognitive}.

Traditional machine learning approaches for tabular medical data, such as gradient boosting methods (XGBoost, LightGBM), typically require hundreds of training samples to achieve clinically acceptable accuracy \cite{moore2022xgboost}. In medical research where each labeled sample involves expensive neuroimaging acquisition, biomarker collection, and longitudinal follow-up, this data requirement creates a significant deployment barrier \cite{ahmed2023barriers}. Moreover, extensive hyperparameter tuning and model development demand additional resources and specialized machine learning expertise that is often unavailable in clinical settings.

Foundation models pre-trained on large-scale synthetic datasets offer a potential solution \cite{woerner2024foundation}. Foundation models have demonstrated promising results across clinical applications, including zero-shot forecasting on physiological time series such as electrocardiogram (ECG) data \cite{zeroshotecg}, suggesting that pre-trained models can transfer effectively to medical prediction tasks without task specific training. Tabular Prior-data Fitted Networks (TabPFN) is a transformer-based model pre-trained on millions of synthetic tabular classification tasks that performs zero-shot learning without task-specific fine-tuning \cite{hollmann2023tabpfn}. Unlike traditional models that learn from scratch, TabPFN leverages pre-trained knowledge through in-context learning, potentially achieving strong performance with minimal training samples. While TabPFN has shown promise on standard benchmarks, its effectiveness for medical prediction tasks with real world constraints—missing values, temporal dependencies, and data leakage risks — remains unexplored. Prior work has demonstrated that machine learning can support clinical decision-making with small medical datasets, for example by fusing imaging and structured clinical data to classify rare breast lesions from limited cohorts \cite{multimodaldeeplearningphyllodes}. Building on this line of research, this study investigates whether pre trained foundation models can match or exceed traditional machine learning performance for MCI-to-AD conversion when training data is severely limited.

\section{Materials and Methods}
\label{sec:methods}

\subsection{Dataset and Preprocessing}
\label{ssec:data}

We used the TADPOLE dataset, derived from ADNI, a longitudinal cohort containing clinical, imaging, and biomarker data from 1,737 participants ~\cite{marinescu2019tadpole}. The dataset includes repeated visits spanning up to 10 years, with structural MRI volumetrics, PET imaging, cognitive assessments, genetic information, and demographic variables collected across time points. We identified participants that had at least one mild cognitive impairment (MCI) visit and excluded 3 participants who had an AD diagnosis at or before their first MCI visit and 396 participants with insufficient follow-up to determine 3-year conversion status. This yielded a final cohort of 530 subjects, of whom 144 converted to AD within 3 years and 386 remained stable. 

The prediction target was conversion from MCI-to-AD within 3 years. For each subject, the first visit with an MCI diagnosis was selected as the baseline time point. Subjects with an AD diagnosis at or before baseline were excluded.  Subjects were labeled as converters if they received an AD diagnosis at any visit within 3 years after baseline. Subjects were labeled as stable non-converters if they had at least one observed follow-up visit at or beyond the 3-year mark with a non-AD diagnosis confirming continued stability. Subjects without sufficient follow-up to determine 3-year status were excluded as ambiguous.

Cognitive test scores, including Mini-Mental State Examination (MMSE), Clinical Dementia Rating Sum of Boxes (CDR-SB), Alzheimer's Disease Assessment Scale-Cognitive Subscale (ADAS-Cog), and Functional Activities Questionnaire (FAQ), were excluded from the feature set to reduce diagnostic overlap and information leakage. Although the TADPOLE dataset contains cerebrospinal fluid (CSF) biomarkers, these were excluded due to high rates of missingness across the cohort~\cite{aracri2025missing}. The final feature set comprised 16 variables: 4 demographic (age, sex, education, APOE $\epsilon$4 status), 7 MRI volumetric (hippocampus, ventricles, whole brain, entorhinal, fusiform, middle temporal, intracranial volume), 3 CSF (A$\beta$, tau, p-tau),  and 2 PET (FDG, AV45) measures.

All preprocessing steps were fit on the training data and then applied to the held-out data. The train/test split was performed at the subject level, ensuring that no visits from the same subject appeared in both sets. Categorical variables such as sex were label encoded using encoders fit on the training set. Missing values were imputed using training set medians. Continuous variables were z-score normalized using training set means and standard deviations. The same preprocessed feature matrix was used for all models to ensure a fair comparison.

\subsection{Model Configuration}
\label{ssec:model}

All experiments used Python 3.12.7 with scikit-learn 1.5.1, xgboost 3.0.2, lightgbm 4.6.0, optuna 4.7.0, and TabPFN 6.3.1, with a fixed random seed of 42 for reproducibility. We evaluated standard tabular machine learning baselines and TabPFN. Logistic regression was fit with a maximum of 1,000 iterations as a linear baseline. Random forest used 200 trees and a maximum depth of 10. LightGBM and XGBoost were configured with 200 estimators, maximum depth 6, and a learning rate of 0.05. A tuned XGBoost variant was additionally evaluated, with hyperparameters selected via Optuna using a Tree-structured Parzen Estimator sampler over 50 trials. The search space included the number of estimators, maximum depth (3--10), learning rate, subsample ratio, column-sample ratio, and L1/L2 regularization parameters. Hyperparameter selection was performed within the training data to prevent test set leakage.

TabPFN was the primary foundation model comparator. We used TabPFN version 6.3.1 (API family V2); V2 was selected as it was the current stable release at the time of experimentation. The implementation in this version is limited to at most 1,000 training samples and 100 features, which did not constrain the present analysis (final cohort N = 530, features = 13). Newer TabPFN releases relax these limits, but our implementation is API-tolerant for both V1 and V2.

\subsection{Experimental Design}
\label{ssec:design}

We conducted three experiments. First, we assessed sample efficiency by comparing models across training set sizes of $N=50$, 100, 200, 500 subjects. For each sample size, we generated 10 stratified training subsets to preserve class proportions and evaluated each fitted model on held-out subjects not used for training.

Second, we evaluated overall model performance on a held-out validation set.The 530-subject cohort was split into a training set of 371 subjects and a held-out test set of 159 subjects using stratified random sampling. The primary metric was AUC, which is threshold-independent and comparatively robust to class imbalance. We additionally report balanced classification accuracy (BCA) ~\cite{grandini2020metrics}, defined as the mean of sensitivity and specificity, to measure classification performance at the default 0.5 decision threshold. AUC differences between TabPFN and each baseline were also assessed using a paired bootstrap test with 1,000 stratified resamples of the test set, with two-sided p-values computed against the null hypothesis of equal AUC.


\begin{figure*}[t]
  \centering
  \includegraphics[width=0.75\textwidth]{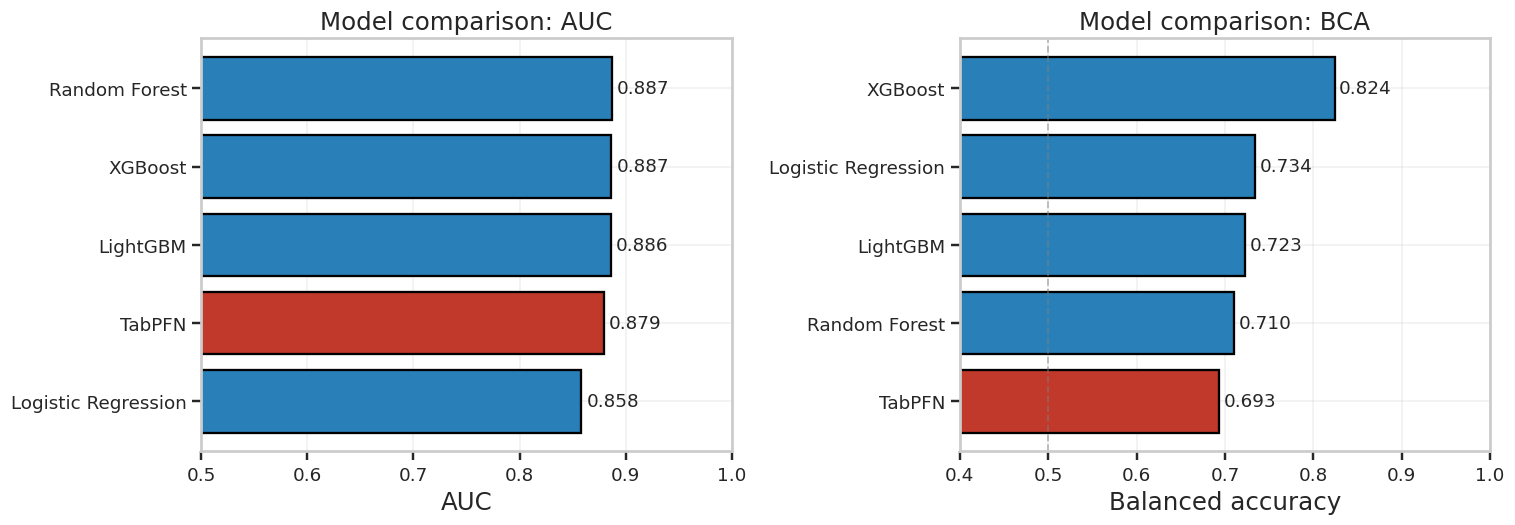}
  \caption{MCI-to-AD Conversion Prediction: Full Model Comparison}
  \label{fig:model_comparison}
\end{figure*}

\begin{figure*}[t]
  \centering
  \includegraphics[width=0.80\textwidth]{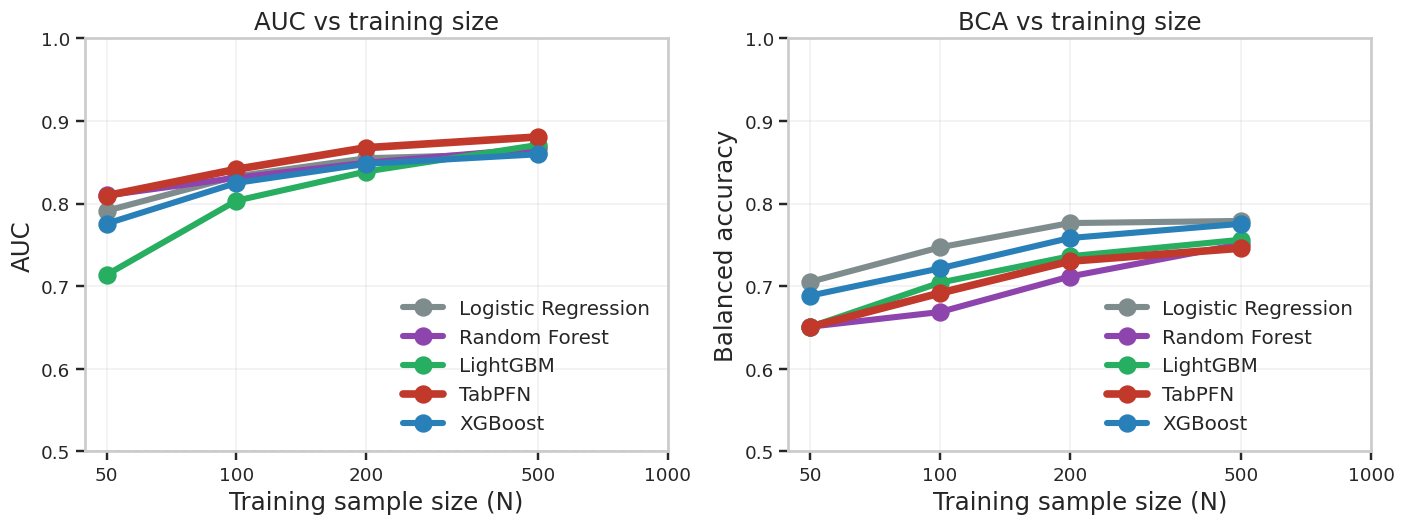}
  \caption{AUC and BCA scores across varying sample sizes}
  \label{fig:sample_sizes}
\end{figure*}

Third, we evaluated the relationship between decision threshold and BCA. The decision threshold is the probability cutoff above which a model classifies a subject as a converter; the default value of 0.5 assumes equal prior probabilities, which may be inappropriate under class imbalance. Because preliminary results showed that TabPFN had high AUC but low BCA at the default threshold, we performed a threshold analysis using cross-validation to avoid test set leakage. For each model, we ran 5-fold stratified cross-validation on the training set, collecting out-of-fold probability estimates for every training patient. The decision threshold was selected as the value that maximized balanced accuracy on these out-of-fold predictions, swept from 0.05 to 0.95 in increments of 0.01. The selected threshold was then applied to the test set predictions.

\section{Results and Discussions}
\label{sec:pagestyle}

\subsection{Full-Data Model Comparison}
Figure ~\ref{fig:model_comparison} presents the overall performance of all models on the held-out test set. Tuned XGBoost and Random Forest tied for the highest AUC at 0.887, with LightGBM close behind at 0.886. TabPFN achieved 0.879, and Logistic Regression performed lowest at 0.858. These results indicate that both tuned gradient boosting and foundation model approaches achieve strong discriminative ability for MCI-to-AD conversion prediction. 

Paired bootstrap tests indicated that the AUC differences between TabPFN and each comparison model were not statistically significant at $\alpha$ = 0.05 (TabPFN vs Logistic Regression: p=0.114; TabPFN vs Random Forest: p=0.618; TabPFN vs LightGBM: p=0.716; TabPFN vs tuned XGBoost: p=0.660). Although the comparison with Logistic Regression approached significance, none of the differences exceeded the conventional threshold, supporting the interpretation that TabPFN achieves competitive ranking performance rather than clear superiority over individual baselines.

Classification performance measured by balanced accuracy revealed a different ranking. XGBoost achieved the highest BCA of 0.824, followed by Logistic Regression at 0.734. LightGBM and the Random Forest configuration achieved 0.723 and 0.710, respectively. TabPFN achieved the lowest BCA of 0.693 despite its strong AUC performance. This discrepancy between AUC and BCA for TabPFN motivated our further threshold analysis.

\begin{figure*}[t]
  \centering
  \includegraphics[width=0.75\textwidth]{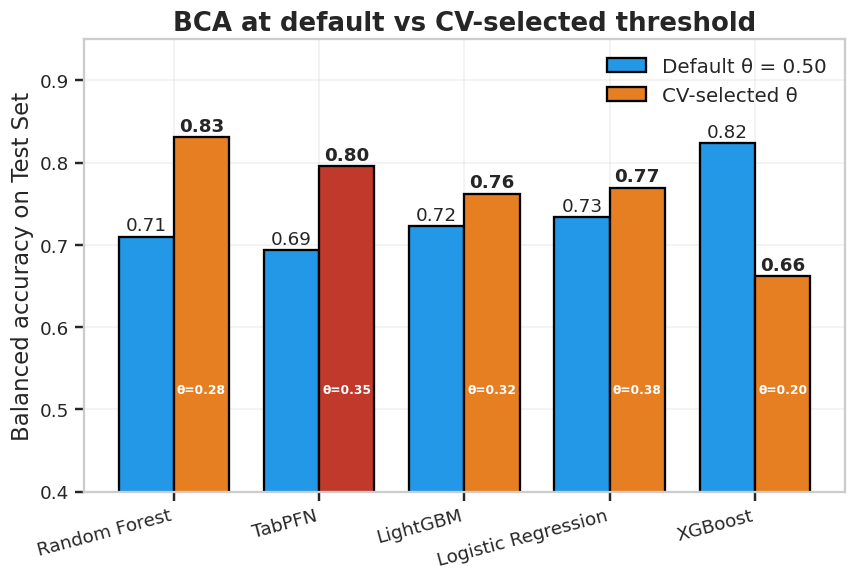}
  \caption{Threshold Optimization Results}
  \label{fig:threshold}
\end{figure*}

\subsection{Sample Efficiency Results}
 Figure ~\ref{fig:sample_sizes} presents model performance across varying training sample sizes. At the smallest training size of N=50, TabPFN achieved the highest AUC of 0.810, outperforming XGBoost (0.776), Logistic Regression (0.791) and LightGBM (0.714) This 9.6 percentage point difference between the best and worst in AUC demonstrates TabPFN's ability to extract predictive signal from limited training data. As training size increased, the performance gap between TabPFN and traditional models narrowed. At N=500, all models converged toward similar AUC values ranging from 0.860 to 0.881, with TabPFN at 0.881 and XGBoost at 0.860. The BCA results across sample sizes revealed a consistent pattern: TabPFN's balanced accuracy lagged behind other models regardless of training size. At N=50, Logistic Regression achieved the highest BCA. The persistent discrepancy between TabPFN's high AUC and low BCA prompted investigation into the model's probability distributions. 

\subsection{Threshold analysis}
Figure ~\ref{fig:threshold} presents BCA at the default and CV-selected thresholds for each model. TabPFN's CV-selected threshold of $\theta$ = 0.35 produced a strong BCA improvement on test data (from 0.69 to 0.80); other models showed mixed results, with Random Forest benefiting most (12\% increase at threshold 0.28). For XGBoost, the CV-selected threshold did not generalize to test data indicating that the model was already approximately well calibrated at $\theta$ = 0.5, and the CV-based selection captured noise rather than systematic miscalibration. TabPFN and Random Forest, by contrast, exhibited consistent benefit from threshold tuning across both training CV and test evaluation. Threshold deviations are not unique to foundation models — both tree-based and foundation model approaches benefited from tuning, while linear and gradient-boosted models did not.

These improvements indicate that the conventional $\theta$ = 0.5 threshold is not always optimal for clinical prediction tasks with class imbalance, and that practitioners should consider cohort-specific threshold selection on a held-out subset rather than relying on the default. From a clinical workflow perspective, lower thresholds correspond to more sensitive screening, which aligns with the clinical preference for detecting MCI-to-AD converters at the cost of additional monitoring for false positives.

\subsection{Clinical Relevance}
Our results demonstrate that TabPFN, a foundation model for tabular data, achieves competitive performance for predicting MCI-to-AD conversion, with particular advantages in data-limited scenarios. At training sizes of 50--100 patients, TabPFN outperformed nearly all traditional machine learning methods, a finding directly relevant to settings where large training cohorts are unavailable — such as early-phase clinical trials, newly established memory clinics, or studies of rare Alzheimer's disease subtypes.

The convergence of model performance at larger sample sizes aligns with theoretical expectations. Traditional gradient boosting methods are highly effective given sufficient training data, and our results confirm that tuned XGBoost achieves strong performance when data availability is not a constraint. The practical implication is that foundation models offer the greatest advantage precisely in the scenarios where they are most needed.

\section{Conclusions}
\label{sec:typestyle}
This study provides a systematic evaluation of TabPFN for predicting MCI-to-AD conversion using biomarker features from the TADPOLE dataset. TabPFN achieved competitive AUC across all training sizes and outperformed traditional machine learning methods at the smallest training sizes (N = 50–100), demonstrating meaningful data efficiency advantages in settings where longitudinal data collection is slow, expensive, and incomplete.

A key practical finding is that TabPFN's strong AUC does not translate directly into strong classification performance at the default 0.5 decision threshold. Threshold optimization improved BCA from 0.69 to 0.80 for TabPFN, and similar gains were observed for other models, indicating that cohort specific threshold selection is important regardless of model choice. Practitioners should not assume that foundation model probability outputs are directly interpretable as risk estimates without validation on representative held-out data.

Several limitations warrant consideration. This evaluation used a single dataset, and generalization to other Alzheimer's cohorts requires further validation. Optimal thresholds are cohort-specific and would require recalibration for different clinical populations. Additionally, while our threshold analysis addresses classification boundary shifts, a formal calibration analysis assessing the Expected Calibration Error (ECE) and utilizing reliability diagrams was not included. Evaluating how well the predicted probabilities map to true empirical likelihoods remains an important step for establishing clinical trustworthiness. The exclusion of cognitive test scores, while methodologically appropriate, limits direct comparison with studies that incorporate wider feature sets.

Future research directions include validation across diverse medical domains to establish whether TabPFN's sample efficiency advantages generalize beyond Alzheimer's disease prediction. Investigation into why foundation models produce systematically conservative probability estimates on clinical data, and whether architectural modifications or pre-training strategies could address this, represents an important avenue for improving clinical applicability. The development of hybrid approaches that combine foundation models for initial risk stratification with traditional methods or clinical expertise for refined assessment may leverage the strengths of both paradigms.

Our findings establish that foundation models represent a practical tool for clinical prediction in data-limited scenarios, provided that appropriate threshold calibration is performed. For settings where large training cohorts are unavailable, TabPFN offers a viable path toward earlier deployment of clinical decision support tools, not as a replacement for traditional methods, but as a complementary approach that excels precisely where data scarcity has historically limited model development.







\bibliographystyle{IEEEbib}
\bibliography{refs}
\end{document}